\documentclass[sigconf]{acmart}
\renewcommand\footnotetextcopyrightpermission[1]{} 
\usepackage{tikz}

\usepackage{pgfplots}
\usepackage{pgfplotstable}
\usepackage{booktabs}
\usepackage{multirow}
\usepackage{makecell}
\usepackage{balance} 
\usepackage{changepage}
\usetikzlibrary{matrix}

\usepackage{endnotes}
\let\footnote=\endnote
 
\pgfplotsset{
	/pgfplots/my xbar legend/.style={
		/pgfplots/legend image code/.code={%
			\draw[##1,/tikz/.cd,bar width=.5em]
			plot coordinates { (\pgfplotbarwidth,0.1em)};}
	}}

\begin{document}
\title[Inferring Missing Categorical Information in Noisy and Sparse
Web Markup]{Inferring Missing Categorical Information \\ in Noisy and Sparse
Web Markup}

\author{Nicolas Tempelmeier, Elena Demidova, Stefan Dietze}

\renewcommand{\shortauthors}{N. Tempelmeier, E. Demidova, S. Dietze}

\affiliation{
  \institution{L3S Research Center, Leibniz Universit\"at Hannover}
  \streetaddress{Appelstr. 9a}
  \city{Hannover}
  \state{Germany}
}
\email{ {tempelmeier, demidova, dietze} @L3S.de}

	\begin{abstract}

Embedded markup of Web pages has seen widespread adoption throughout the past 
years driven by standards such as RDFa and Microdata and initiatives such as schema.org, where recent studies show an adoption by 39\% of all Web pages already in 2016. While this constitutes an important information source for tasks such as Web search, Web page classification or knowledge graph augmentation, individual markup nodes are usually sparsely described 
and often lack essential information. For instance, from 26 million nodes describing events within the Common Crawl in 2016, 59\% of nodes provide less 
than six statements and only 257,000 nodes (0.96\%) are typed with more specific event subtypes.
Nevertheless, given the scale and diversity of Web markup data, nodes that provide missing information can be obtained from the Web in large quantities, in particular for categorical properties. Such data constitutes potential training data for inferring missing information to significantly augment sparsely described nodes. 
In this work, we introduce a supervised approach for inferring missing categorical properties in Web markup. Our experiments, conducted on properties of events and movies, show a performance of 79\% and 83\% F1 score correspondingly, significantly outperforming existing baselines.

\end{abstract}

	\maketitle

		\thispagestyle{empty}

	\section{Introduction}
\label{sec:introduction}

Web search in general and the interpretation of Web documents in particular are increasingly being supported through semi-structured, entity-centric knowledge. 
For instance, publicly available knowledge graphs (KGs) such as Freebase~\cite{Bollacker:2008:FCC:1376616.1376746} or YAGO~\cite{suchanek2007yago} as well as proprietary KGs used by Google or Microsoft \cite{googleGraph, paulheim2016knowledge} are key ingredients when interpreting search queries as well as Web documents. 
More recently, Web markup facilitated through standards such as RDFa\footnote{RDFa W3C recommendation: \url{http://www.w3.org/TR/xhtml-rdfa-primer/}}, Microdata\footnote{\url{https://www.w3.org/TR/microdata/}} and Microformats\footnote{\url{http://microformats.org}} has become prevalent on the Web, driven by initiatives such as \emph{schema.org}, 
a joint effort led by Google, Yahoo!, Bing and Yandex. 

For instance, 
the Web Data Commons (WDC) project \cite{Meusel:2014:WMR:2717213.2717235} that
releases markup extracted from the 
Common Crawl\footnote{\url{http://commoncrawl.org/}},
found that in 2016 39\% out of 3.18 billion HTML pages from over 34 million 
pay-level-domains (plds) contain some form of embedded markup, resulting in 
a corpus of 44.24 billion 
RDF quadruples\footnote{\url{http://webdatacommons.org/structureddata/2016-10/stats/stats.html}}. 
There is an upward trend of Web markup adoption, where the proportion of
pages containing markup increased from 5.76\% to 39\% between 2010 and 2016. 

To this extent, markup data provides an unprecedented and growing source of explicit entity annotations to be used when interpreting and retrieving Web documents, to complement annotations otherwise obtainable through traditional information extraction 
pipelines, or to train information extraction methods.
In addition, while traditional KGs capture large amounts of factual knowledge, 
they still are incomplete, i.e. coverage and completeness vary heavily across 
different types or domains. In particular, there is a large percentage of less 
popular (long-tail) entities and properties that are usually insufficiently represented \cite{Bizer2013}.
In this context, markup also provides essential input when incrementally 
augmenting and maintaining KGs \cite{conf/icde/YuGFD17}, in particular when
attempting to complement information about long-tail properties and entities
\cite{yu2018knowmore}.

The specific characteristics of statements extracted from embedded Web markup
pose particular challenges~\cite{yu2016iswc}. 
Whereas coreferences are very frequent (for instance, in the WDC 2013 corpus, 
18,000 entity descriptions of type \textit{schema.org:Product} are returned for the 
query \texttt{`Iphone 6'}), these are not linked through explicit statements. In contrast to traditional densely connected RDF graphs, markup statements mostly consist of isolated nodes and small subgraphs, each usually made up of small sets of statements per entity description. 
In addition, extracted RDF markup statements are highly redundant 
and are often limited to a small set of highly popular predicates, such as
\textit{schema.org:name}, complemented by a long tail of less frequent
statements. Moreover, data extracted from markup contains a wide variety of errors~\cite{meusel2016towards}, ranging from typos to the frequent
misuse of vocabulary terms~\cite{conf/esws/MeuselP15}.
Hence, individual markup extracted from a particular Web document or crawl 
usually contains very limited or unreliable information about a particular
entity. 
According to our analysis, out of 26 million annotated events in the WDC 2016
corpus, less than 257,000 (0.96\%) indicate a more specific event subtype and 59\% nodes provide less than six statements.
This strongly limits the meaningfulness of Web markup, in particular for
entities that cannot be mapped to a representation in an existing knowledge
graph. 

In this work, we introduce an approach to automatically infer missing
categorical information for particular entities obtained from Web markup.
Building on the Web-scale availability of markup, and hence, 
the abundance of potential training data for the task, we introduce a supervised method to efficiently infer missing categorical information 
from existing entity markup describing coreferring or similar entities. 
Our experiments address the inference of entity (sub-)types, as well as inference of arbitrary non-hierarchical predicates, such as movie genres. 
We demonstrate superior performance compared to both naive baselines 
as well as specialised state-of-the-art methods for type inference and achieve F1 scores of 79\% and 83\% in two experimental tasks.

	\section{Motivation and Problem Definition}
\label{sec:problem}
\subsection{Motivation}

Microdata and RDFa markup are used to embed semi-structured data about entities within Websites. While being leveraged to facilitate interpretation and retrieval 
of Websites by most 
major search engines, markup data is also used to maintain and augment knowledge graphs, 
where additional applications include Google Rich Snippets, Pinterest Rich Pins 
and search features for Apple Siri \cite{Guha:2016:SES:2886013.2844544}. 
By today, Web markup data is available at an unprecedentedly large scale, which can 
be exemplarily observed on the \emph{Web Data Commons}
(WDC)~\cite{Meusel:2014:WMR:2717213.2717235} corpus. 
WDC\footnote{\url{http://webdatacommons.org/}} offers a large-scale 
corpus of RDF quadruples extracted from the \emph{Common Crawl}\footnote{\url{http://commoncrawl.org/}}. 
The crawl of October 2016 contains $3.18 \cdot 10^9$ URLs of which 39\% exhibit
markup from which over $4.4 \cdot 10^{10}$ triples were extracted. 
In contrast, the crawl of November 2015 contains $~1.77 \cdot 10^9$ URLs of which
only 31\% exhibit markup, resulting in only 
about  $2.4 \cdot 10^{10}$ extracted 
triples\footnote{Detailed numbers can be found at
\url{http://webdatacommons.org/structureddata/index.html}}. \emph{Schema.org} is a joint initiative from major search engines such as 
Bing, Google, Yahoo! and Yandex that provides a joint vocabulary and is the most
commonly deployed vocabulary on the Web \cite{Bizer2013}.
In the following we abbreviate the prefix 
of the \emph{schema.org} vocabulary by \emph{s:}, e.g. \emph{s:Movie}. 
We refer to the WDC corpus from October 2016 
as the WDC 2016 corpus.

\begin{table*}
	\caption{Number of quadruples per node for specific types in WDC 2016.}
	\begin{tabular}{rrrrrrrrrrr}
		\toprule

		\multirow{2}{*}{\raisebox{-\heavyrulewidth}{\textbf{Type}}} & 
		\multirow{2}{*}{\raisebox{-\heavyrulewidth}{\textbf{\makecell[c]{Total No.\\Quadruples}}}} &
		\multirow{2}{*}{\raisebox{-\heavyrulewidth}{\textbf{\makecell[c]{Total No.\\Nodes}}}} &
			\multicolumn{4}{c}{\textbf{Quadruples}} &
			\multicolumn{4}{c}{\textbf{Distinct Properties}}\\
			\cmidrule(l{4pt}r{4pt}){4-7}
			\cmidrule(l{4pt}r{4pt}){8-11}
			
		&  &  & \textbf{Min.} & \textbf{Max} & \textbf{Avg.} & \textbf{Median}
						& \textbf{Min.} & \textbf{Max.} & \textbf{Avg.} & \textbf{Median}\\
			
		\midrule
		
		\emph{s:Event} & $1.58 \cdot 10^8$ & $2.66 \cdot 10^7$ & 1& 2889 & 5.55 & 5 & 1 & 32 & 5.31 & 5 \\
		\emph{s:Movie} & $1.25 \cdot 10^8$ &  $1.62 \cdot 10^7 $ & 1 & 4547 & 7.71 & 6 & 1 & 26 & 5.77 & 6 \\
		\bottomrule
	\end{tabular}
	\label{prob:dataSetTable}
\end{table*}

While Web markup constitutes an unprecedented source of semi-structured
knowledge, markup is usually sparse and highly redundant, consisting of 
vast amounts of coreferences and (near) 
duplicate statements \cite{conf/icde/YuGFD17}. 
Individual entities extracted from Web markup usually are sparsely described,
such that only a fraction of the properties foreseen by \emph{schema.org} for a
specific type is provided, often only providing a label and a type for a
specific node. Table \ref{prob:dataSetTable} provides an overview 
of the number of quadruples per single node for specific types (\emph{s:Event}, \emph{s:Movie}) in the WDC 2016 corpus.
The property distribution follows a power law, where a small set of terms is very prevalent, yet the majority of properties is hardly used across the Web. 
Figure \ref{probDef:MovieTopProps} shows the top-20 most frequently used 
properties of movies, highlighting that certain properties occur very often 
(e.g. \emph{s:actor}) while others are provided rarely, 
such as \emph{s:productionCompany}. \par
Sparsity is exacerbated by the lack of connectivity of markup data, 
where controlled vocabularies, taxonomies, and essentially, 
links among nodes are hardly present. 
Previous studies \cite{Dietze:2017:AIE:3041021.3054160} on a specific markup 
subset find that, out of a set of 46 million quadruples involving transversal, 
i.e. non-hierarchical properties, approximately 97\% actually refer to literals 
rather than URIs, that is object nodes. These findings underline that markup data 
largely consists of rather isolated nodes, which are linked through 
common schema terms (as provided by \emph{schema.org}) at best, but commonly lack relations at the instance level. In particular for categorical information, such as movie genres or product categories, this poses a crucial challenge when it comes to 
interpreting such information.

\pgfplotstableread[col sep=comma,]{figures/top20Movies.csv}\topMoviesTab
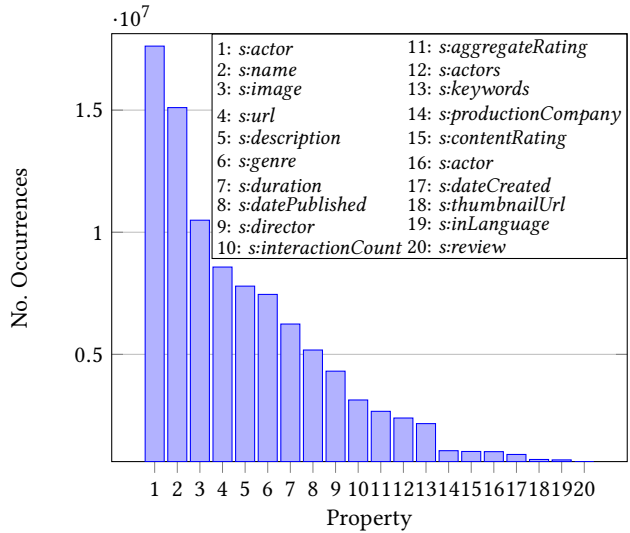
\begin{figure}
	\begin{tikzpicture}
	\begin{axis}[ybar,
	xtick=data,
	xticklabels from table={\topMoviesTab}{rowNum},
	xlabel={Property},
	ylabel={No. Occurrences},
	ymajorgrids,
	xtick pos=left,
	enlarge y limits={0.03,upper},    
	bar width=0.8em]
	\addplot table[x expr=\coordindex, y={occurrences}]{\topMoviesTab};
	\end{axis}
	
	\matrix[
	matrix of nodes,
	anchor=north east,
	draw,
	fill=white,
	inner sep=0.1em,
	column 1/.style={nodes={anchor=west, font=\small}},
	column 2/.style={nodes={anchor=west, font=\small}},
	draw
	]
	at([xshift=0cm]current axis.north east){
		1: \textit{s:actor} & 11: \textit{s:aggregateRating}\\
		2: \textit{s:name} & 12: \textit{s:actors} \\
		3: \textit{s:image}& 13: \textit{s:keywords}\\
		4: \textit{s:url} & 14: \textit{s:productionCompany} \\
		5: \textit{s:description} & 15: \textit{s:contentRating}\\
		6: \textit{s:genre} & 16: \textit{s:actor} \\
		7: \textit{s:duration} & 17: \textit{s:dateCreated}\\
		8: \textit{s:datePublished} & 18: \textit{s:thumbnailUrl}\\
		9: \textit{s:director} & 19: \textit{s:inLanguage}\\
		10: \textit{s:interactionCount}& 20: \textit{s:review}\\
		};
	\end{tikzpicture}
	\caption{Top-20 most frequent properties for the type \emph{s:Movie} in WDC 2016. The second entry of \textit{s:actor} is caused by erroneous annotations in Web markup.}
	\label{probDef:MovieTopProps}
	
\end{figure}

A particular instantiation of the aforementioned problem is the use of unspecific 
types. Figure \ref{prob:occEvents} illustrates the number of instances of events 
annotated with respective event subtypes. Note that assignment of multiple types 
is theoretically possible, but rarely used in practice (i.e. less than 0.1\% of 
events have multiple types). Apparently, most of the instances are assigned the 
generic type \emph{s:Event}, while only 0.96\% of nodes use more specific types 
like \emph{s:TheaterEvent} or \emph{s:Festival}, hindering data interpretation.

\pgfplotstableread[col sep=comma,]{figures/eventTypeDist.csv}\evDistTab
\begin{figure}
	\begin{tikzpicture}
	\begin{axis}[ybar,
	xtick=data,
	xticklabels from table={\evDistTab}{rowNum},
    ylabel={No. Occurrences},
    xlabel={Type},     
    ymode=log,
    xtick pos=left,
    ymajorgrids,
	enlarge y limits={0.03,upper},    
	bar width=0.8em]
		\addplot table[x expr=\coordindex, y={count}]{\evDistTab};
	\end{axis}

		\matrix[
		matrix of nodes,
		anchor=north east,
		draw,
		fill=white,
		inner sep=0.1em,
		column 1/.style={nodes={anchor=west, font=\small}},
		column 2/.style={nodes={anchor=west, font=\small}},
		draw
		]
		at([xshift=0cm]current axis.north east){
			1: \textit{s:Event} & 11: \textit{s:FoodEvent}\\
			2: \textit{s:PublicationEvent} & 12: \textit{s:DanceEvent} \\
			3: \textit{s:MusicEvent}& 13: \textit{s:SocialEvent}\\
			4: \textit{s:ScreeningEvent} & 14: \textit{s:ChildrensEvent}\\
			5: \textit{s:EducationEvent} & 15: \textit{s:Festival} \\
			6: \textit{s:VisualArtsEvent} & 16: \textit{s:SaleEvent}\\
			7: \textit{s:TheaterEvent} & 17: \textit{s:ExhibitionEvent}\\
			8: \textit{s:ComedyEvent} & 18: \textit{s:BusinessEvent}\\
			9: \textit{s:SportsEvent} & \\
			10: \textit{s:LiteraryEvent}& \\
		};

	\end{tikzpicture}
	\caption{Number of occurrences of schema.org event types in WDC 2016 (Y-axis is logarithmic).}
	\label{prob:occEvents}
\end{figure}
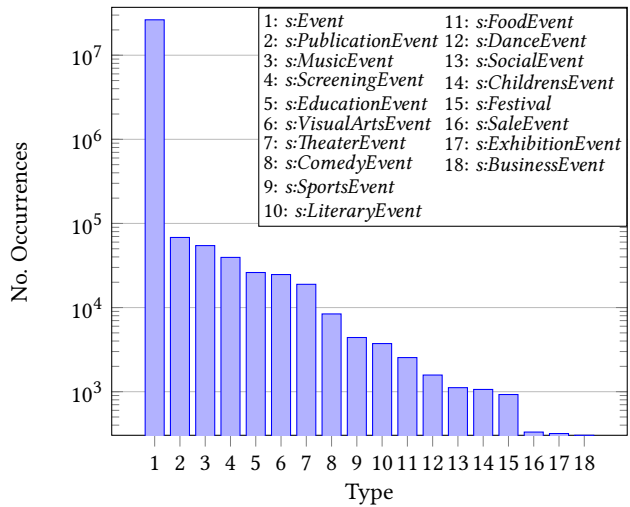

Whereas individual markup nodes are usually sparsely annotated, markup as a whole 
provides a rich source of data, where in particular for categorical, i.e. discrete, 
properties a wide variety of instances can be drawn from the long tail. 
For instance, referring to Figure \ref{prob:occEvents}, while only 0.96\% of 
all event nodes are typed with a meaningful subtype, this still corresponds to 
a set of 257,000 nodes available as training data to build supervised models to 
classify the remaining 26 million insufficiently typed events.
Hence, we follow the intuition that markup data can significantly benefit from 
supervised approaches, which learn categorial or discretised properties as a means 
to infer missing categorical information for sparsely annotated nodes, i.e. to 
enrich markup entities. 
Overall, augmentation of sparse Web markup nodes can contribute to the 
improvement of the interpretability of the markup, the enrichment of knowledge 
graphs, and hence, to the effectiveness of the applications using the markup. 
This includes search and Web page classification, where in particular categorical 
and type information is essential to correctly interpret resources.

\subsection{Problem Definition}

This work aims at inferring missing categorical information in data sourced from Web markup.
For a given corpus of Websites $\mathcal{C}$, $\mathcal{Q}_\mathcal{C}$ denotes the set 
of \emph{RDF quadruples} of the form $(s,p,o,u)$ extracted from the corpus, where $s,p,o$ represent 
an RDF triple, i.e. a statement, of the form subject, predicate and object and $u$ represents the URL of the Web document, 
from which the triple has been extracted. 

A \emph{vocabulary} $V$ consists of a set of \emph{types} $T$ and
\emph{properties} $P$.
A particular property $p_i \in P$ has a declared domain $d(p_i)$ that defines the set of expected 
types $T_i \subseteq T$ a subject involved in the same triple with $p_i$ is
meant to be an instance of.
The range $r(p_i)$ of a property $p_i$ defines the expected types an object involved in the 
same triple as $p_i$ is meant to be an instance of.

For instance, within the \emph{schema.org} vocabulary, the domain of 
the property \emph{translator}\footnote{\url{http://schema.org/translator}} is defined 
as instances of type \emph{Event}\footnote{\url{http://schema.org/Event}} 
and \emph{CreativeWork}\footnote{\url{http://schema.org/CreativeWork}}, while the
declared range is defined as instances of type \emph{Organization}\footnote{\url{http://schema.org/Organization}} 
and \emph{Person}\footnote{\url{http://schema.org/Person}}.

\begin{definition} 
Given a vocabulary $V$, a set of quadruples $\mathcal{Q}_\mathcal{C}$, 
for a particular node representing a subject $s_i \in \mathcal{Q}_\mathcal{C}$, this
work aims at predicting quadruples $q=(s_i,p_i,o_i,u_i)$ which are: (a) not present in the markup corpus ($q \notin \mathcal{Q}_\mathcal{C}$), (b) valid according to the definition of vocabulary $V$, and (c) a valid statement about subject $s_i$ in the context of $u_i$.
\end{definition}

The last requirement of the aforementioned definition is experimentally evaluated according to a ground truth $G$, where an example is described in Section \ref{sec:dataset}.

Note that our work focuses on \emph{categorical} properties, 
i.e. we consider properties where the corresponding range $r(p)$ is \emph{finite}. 

For instance, consider the following markup triple, extracted 
from the URL \url{http://www.imdb.com/title/tt0109830/?ref\_=tt\_trv\_cnn} 
describing the movie "Forrest Gump":
\[\left[
\begin{array}{l|l}
	s: & \emph{\_:nodea73846c741abe988abf1c682f1fe26e7}\\
	p: & \emph{rdf:type}\\
	o: & \emph{s:Movie}\\
\end{array}\\[0.5em]
\right]\]
For the specific subtask of predicting movie genres (Section \ref{sec:evaluation}), we aim at predicting the quadruple involving the following triple (URL omitted) stating the genre of the movie:
\[\left[
\begin{array}{l|l}
	s: & \emph{\_:nodea73846c741abe988abf1c682f1fe26e7}\\
	p: & \emph{s:genre}\\
	o: & "Drama"\\	
\end{array}
\right]\]

\section{Approach}
\label{sec:architecture}

The characteristics of the data at hand suggest that, for most subjects $s_i$ which are to be augmented, e.g. the movie mentioned in the previous example, sufficient training data can be obtained (Section \ref{sec:problem}). That means, we anticipate that a sufficient number of entity descriptions (instances) exist, which share the same missing categorical property $p_i$, e.g. a movie genre in the example above. Thus, we approach the inference problem as a supervised classification problem, where nodes which share the sought after property $p_i$ are used as training data to build a model for the prediction of respective statements. 
This section describes our approach, namely the steps taken for data cleansing, feature extraction and building classification models.    

\subsection{Data Cleansing}
Based on studies on common errors on deployed microdata 
\cite{conf/esws/MeuselP15},
we applied the following heuristics proposed in 
\cite{conf/esws/MeuselP15},
to improve the quality 
of the dataset by fixing the following errors:

	\textbf{Wrong namespaces}: Many terms that deviate from the correct
	\emph{schema.org} namespace can be corrected by adding missing slashes, changing \textit{https://} to \textit{http://}, removing additional substrings between \textit{http://} and \textit{schema.org} and fixing capitalisation errors.

	\textbf{Undefined properties and types}: 
	The use of wrong capitalisation of property and type names leads to the presence 
	of undefined terms in markup data. 
	We corrected the capitalisation by using the capitalisation 
	defined by the \emph{schema.org} vocabulary.   

Applying these heuristics aids the feature extraction and classification steps described below by providing
a larger amount of training data as well as by improving feature quality.

\subsection{Feature Extraction}

This section describes the considered features for our task and the applied feature extraction.

\textbf{\emph{pld/tld}}: Based on the assumption that many Web domains are specialised on particular topics, e.g. concerts or documentary films, we employ domain-based features. The intuition is that any particular \emph{pay-level-domain} (pld) and/or \emph{top-level-domain} (tld) usually correlates with particular categorical properties, such as the types of covered events.
Thus, for each node, we extract the pld and the tld from the URL of the Web page. 
For instance, taking into account the task of predicting event subtypes, 
consider the quadruple:
\[\left[
\begin{array}{l|l}
	s:& \emph{\_:node396540c21b6fa0388c7293ebe216583}\\
	p:& \emph{rdf:type}\\
	o:& \emph{s:Event}\\
	u: & \emph{<http://www.touristlink.com/india/cat/events.html>}\\
\end{array}\\[0.5em]
\right]\]

\noindent From this quadruple we extract the pld "touristlink.com" and the tld
".com" from $u$ and use these as features to predict 
the subtype "\emph{s:MusicEvent}" of the described event.
The plds and tlds are mapped into feature space via \emph{1-hot-encoding}\footnote{For a brief description of 1-hot-encoding see: \url{http://scikit-learn.org/stable/modules/preprocessing.html\#preprocessing-categorical-features}},
resulting in one dimension for each pld and each tld.

\textbf{\emph{node-vocab}}: 
The intuition behind this feature is that there is a correlation between the used vocabulary terms and the specific classes we aim to predict. For example, a composer (\emph{s:composer}) is more likely to be provided for a music event (\emph{s:MusicEvent}) than for a sports event. Following this intuition, Paulheim et al. \cite{Paulheim:2013}
proposed an approach for entity type prediction using vocabulary term correlations. To this extent, they made use of the outgoing and incoming statements of the node $n$ for type prediction of $n$ in knowledge graphs (i.e. statements that have $n$ either in the subject or the object position, respectively). In case of Web markup, it may not be feasible to determine all incoming statements for a given subject at Web scale. Therefore, in this work we make use of the outgoing statements only and use these statements to predict categorical properties of the entity described through the node $n$.
More specifically, for all quadruples $Q_n$ involving subject $n$, we extract all \emph{schema.org} terms used as predicate.
For each node $n$, we compute a frequency vector, where each dimension corresponds to a vocabulary term $t_i$ and each value is the normalised number of times $t_i$ occurs in a quadruple with $n$ as a subject. The frequencies are normalised using the $l^2$ (euclidean) norm.
\begin{example}
\label{arch:ExampleNode}
For the node $s$ and URL $u$

\[\left[
\begin{array}{l|l}
	$s$:& \emph{\_:node3957c770b4f7c0bd1a17805dd8ca406}\\
	$u$:& \emph{<https://gdssummits.com/nghealthcare/us/>}\\
\end{array}
\right]\]
the following tuples are present:
\begin{align*}
&\left[
\begin{array}{l|l}
	$p$:& \emph{rdf:type}\\
	$o$:& \emph{<http://schema.org/BusinessEvent>}\\
\end{array}
\right]\\
&\left[
\begin{array}{l|l}
	$p$:& \emph{s:Event/name}\\
	$o$:& \emph{"NG Healthcare Summit US"@en}\\
\end{array}
\right]\\
&\left[
\begin{array}{l|l}
	$p$:& \emph{s:Event/location}\\
	$o$:& \emph{"Omni Barton Creek Resort \& Spa, Austin, Texas"@en}\\
\end{array}
\right]
\end{align*}

These tuples result in the following node-vocab: \{\textit{rdf:type}:1,\\
\textit{s:Event/name}:1, \textit{s:Event/location}:1\}.
\end{example}
Note that we concatenated the predicate and the type used as the
domain of the predicate. This way we ensure that: (a) 
types as well as terms are considered and (b) the connection between a 
predicate and its observed domain is preserved. 
The latter appears useful, 
considering that \emph{schema.org} terms are used in a variety of contexts, 
often in ways other than recommended by the vocabulary definition, 
e.g. by violating domain and range 
definitions~\cite{Dietze:2017:AIE:3041021.3054160}.

\textbf{\emph{page-vocab}}: 
The vocabulary used on a Web page within which a subject appears intuitively correlates with categorical classes associated with nodes on the respective page. 
For instance, Websites discussing music albums are more likely to also contain
music events rather than sports events. To take this context into account, we consider all \emph{schema.org} vocabulary terms that appear as predicates on the same Web page as the node under consideration as a feature.
Similar to the node-vocab, we create a frequency vector normalised using the $l^2$ (euclidean) norm.

\begin{example}
	Assume that in addition to the quadruples in 
	Example~\ref{arch:ExampleNode}, the following triples are present on the
	same Web page:
\begin{align*}
	&\left[
	\begin{array}{l|l}
		$s$ & \emph{\_:nodea9ff152514bcfb63c2714bc1336b2b3} \\
		$p$:& \emph{s:Organization/url}\\
		$o$:& \emph{<http://www.gdsinternational.com>}\\
	\end{array}
	\right]\\
	&\left[
	\begin{array}{l|l}
		$s$ & \emph{\_:node4ccbf7f34c95f14168f5fdb47b73ab} \\
		$p$:& \emph{rdf:type}\\
		$o$:& \emph{s:BusinessEvent}\\
	\end{array}
	\right]
\end{align*}
Then the terms from these quadruples are added to the node-vocab to form the
page-vocab: \{\textit{rdf:type}:2, \textit{s:Event/name}:1, \textit{s:Event/location}:1, \textit{s:Organization/url}:1\}.

\end{example}

After computing the individual features, all features are concatenated to form a single 
feature vector. Finally, the feature vectors are normalised, i.e. the mean is removed and the features are scaled to unit variance. The feature vectors serve as input for supervised machine learning approaches that are detailed in Section
\ref{sec:ClassTuning}.

\subsection{Classification Models}
\label{sec:ClassTuning}
We compare the use of the following classifiers: 

	{\sc \bf Na\"ive Bayes}: A Gaussian Na\"ive Bayes classifier 
	that assumes that the likelihood of the 
	features follows a Gaussian distribution. Since the features are normalised
	(i.e. may have negative values), a multinomial Na\"ive Bayes can not be applied.
	Na\"ive Bayes classifiers are known to be adoptable to many classification tasks.
	
	{\sc \bf Decision Tree}: A classifier that successively divides the
	feature space to maximise a given metric (e.g. Gini Impurity, Information Gain). 
	Decision Trees are able to identify discriminative 
	features within high-dimensional data.
	
	{\sc \bf Random Forest}: A classifier that utilises an ensemble of
	uncorrelated decision trees. Random Forests can utilise a large amount 
	of training data that is likely to be found in Web crawls.
	
	{\sc \bf SVM}: A Support Vector Machine with a linear kernel. SVMs have
	been applied to a large variety of classification problems.

	\section{Evaluation Setup}
\label{sec:evaluation}
While our approach is independent of the respective categorical information to be inferred, we conducted an evaluation in two specific tasks: (1) predicting subtypes of \textit{s:Event} instances, and (2) predicting genres (\textit{s:genre}) of \textit{s:Movie} instances.

\subsection{Datasets}
\label{sec:dataset}

Training and test datasets were extracted from the Web Data Commons dataset of October 2016.

\textbf{Event Classification}: This task deals with the prediction of 
event subtypes. \emph{Schema.org} distinguishes between 19 
different event subtypes, such as \emph{s:BusinessEvent} or \emph{s:SportsEvent}. 
Given a generic event, the goal of this task is to predict the correct subtype 
of the event, i.e. to predict the object of the
\emph{rdf:type} statement.

\textbf{Movie Genre Classification}: \emph{Schema.org} allows annotation of movie genres via the \emph{s:genre} property. The goal of this classification task is to predict statements describing the \emph{s:genre} of respective movies. 
Since it is possible to assign multiple genres to a single movie by 
defining multiple \emph{s:genre} properties, the classification of movie genres is a \emph{multi-label problem}, i.e. a single movie entity can belong to multiple genre classes. We address this multi-label problem by extracting individual datasets for each genre upon which a binary classifier for each genre is trained.

\pgfplotstableread[col sep=comma,]{figures/visualarts.csv}\visArtDist
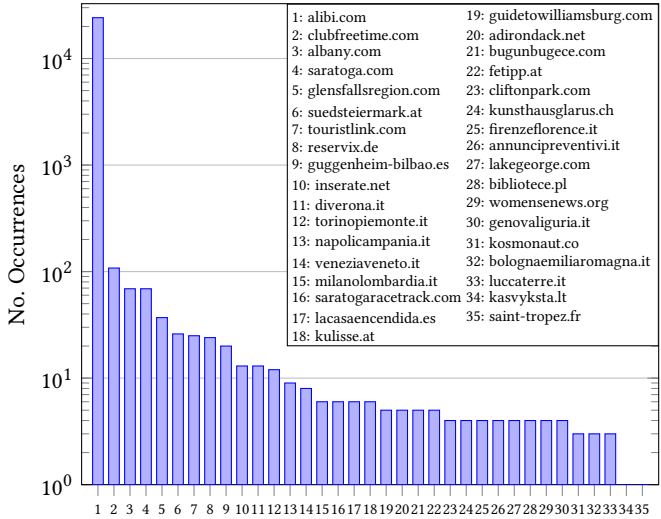
\begin{figure}
	\begin{tikzpicture}
	\begin{axis}[ybar,
	xtick=data,
	xticklabels from table={\visArtDist}{rowNum},
	xticklabel style={font=\scriptsize},
	ylabel={No. Occurrences},            
	ymode=log,
	xtick pos=left,
	enlarge x limits={0.03},
	ylabel near ticks,
	ymajorgrids,
	width=0.52\textwidth,
	enlarge y limits={0.03,upper},
	bar width=0.45em]
	\addplot table[x expr=\coordindex, y={count}]{\visArtDist};
	\end{axis}

		\matrix[
		matrix of nodes,
		anchor=north east,
		draw,
		fill=white,
		inner sep=0.1em,
		column 1/.style={nodes={anchor=west, font=\scriptsize}},
		column 2/.style={nodes={anchor=west, font=\scriptsize}},
		draw
		]
		at([xshift=0cm]current axis.north east){
			1: alibi.com & 19: guidetowilliamsburg.com \\
			2: clubfreetime.com & 20: adirondack.net \\
			3: albany.com & 21: bugunbugece.com \\
			4: saratoga.com & 22: fetipp.at \\
			5: glensfallsregion.com & 23: cliftonpark.com \\
			6: suedsteiermark.at & 24: kunsthausglarus.ch \\
			7: touristlink.com  & 25: firenzeflorence.it \\
			8: reservix.de & 26: annuncipreventivi.it \\
			9: guggenheim-bilbao.es & 27: lakegeorge.com \\
			10: inserate.net & 28: bibliotece.pl \\
			11: diverona.it & 29: womensenews.org \\
			12: torinopiemonte.it  & 30: genovaliguria.it \\
			13: napolicampania.it & 31: kosmonaut.co \\
			14: veneziaveneto.it & 32: bolognaemiliaromagna.it \\
			15: milanolombardia.it & 33: luccaterre.it \\
			16: saratogaracetrack.com & 34: kasvyksta.lt \\
			17: lacasaencendida.es & 35: saint-tropez.fr\\
			18: kulisse.at &\\
		};

	\end{tikzpicture}
	\caption{tld/pld-distribution of \emph{s:VisualArtsEvent}s. Y-axis is
		logarithmic.}
	\label{ev:longTail}
\end{figure}

\subsubsection{Balancing and Sampling}

We extracted quadruples that exhibit the respective property of interest by selecting quadruples which describe nodes of \emph{rdf:type} \emph{s:Event} (\emph{s:Movie}) and are annotated with a more specific event subtype in the case of events and the 
\emph{s:genre} predicate for movies. This results into a 
single $Events$ dataset (containing instances of all considered  subtypes) and an individual dataset for each movie genre.
As illustrated in Figure \ref{prob:occEvents}, the class distribution is uneven.

To obtain a balanced dataset that is sufficiently large for training of a machine learning algorithm, we applied the following steps. For $Events$, we picked the top-7 classes with the highest number of instances. We introduced an additional class containing all events not included in the top-7 classes. 
The classes were balanced by limiting the size of all classes to $c_e$, which is the size of the smallest class.  
For $Movies$, we extracted 7 individual datasets corresponding to the top-7 most frequent movie genres. Each individual genre dataset includes all instances of the particular genre as well as all the remaining instances, which are labeled as "Other". The size of each genre datasets is limited to $c_m$, which is the size of the smallest class among all 7 datasets.

We employed two different sampling strategies: 

1) \emph{Stratified Random Sampling} 
simply chooses $c_e$ ($c_m$) instances of each class at random from the whole dataset. 

2) \emph{pld-Aware Sampling}: Figure
\ref{ev:longTail} depicts the pld distribution of \emph{s:VisualArtsEvent}s. 
The distribution follows a power law, such that a small set of plds provides the majority of events. 
Random sampling may result in dropping some of the plds 
with fewer events and overfitting towards the patterns exhibited by very prominent plds. Therefore, we employ a sampling approach that ensures representation of long-tail entities in the sample. To this extent, we calculate a \emph{fair share} in the sample by dividing the number of instances by the numbers of plds. We add all instances 
from plds that have fewer instances than the \emph{fair share}. 
This process is repeated with recalculating the \emph{fair share} with respect to the number of missing instances until the dataset contains $c_e$ ($c_m$) instances of each class, where $c_e$ ($c_m$) is the number of instances of the smallest class in the case of events (movies). If all remaining plds contain more instances than the \emph{fair share}, each pld contributes the \emph{fair share} to the final sample. 

After the sampling, we split each resulting dataset 
in an individual training and test set (80\% / 20\% of the instances).

\subsubsection{Labeling \& Ground Truth}

We follow a dataset-specific strategy to obtain class labels, 
i.e. a ground truth for training and testing. 
For assigning event types, we rely on the event subtypes 
defined within the \emph{schema.org} type hierarchy.
The class labels for events are thus explicitly given by the
\textit{rdf:type}-statements.

With respect to the prediction of movie genres, no controlled vocabulary is used consistently, whereas literals are used widely. Therefore, we map the literals to a unified genre taxonomy. We make use of the 22 genres defined by the International Movie Database (IMDB)\footnote{\url{http://www.imdb.com/genre/}}. To obtain the class labels, we check for string containment of the IMDB genre names in the literal values of the \emph{s:genre} properties. If a genre name is a substring of the aforementioned property the genre is assigned 
as class label to the respective instance. Note that it is possible for one instance to exhibit multiple labels since multiple genre names may be substrings of a single \emph{s:genre} property and, in addition, single instances may have multiple \emph{s:genre} properties. 
Intuitively, this process leads to reasonable class labels for the majority of instances, such that a sufficiently large amount of correctly labeled training data can be obtained. Yet, we also anticipate a certain amount of noise. The cleansed and labeled datasets are made publicly available\footnote{The datasets can be found at \url{http://markup.l3s.de}.}.

Table \ref{dat:DatasetNotation} provides an overview of the size of the extracted datasets as well as the amount of included plds. 
The event datasets are denoted by $Events$ and contain the following classes: \textit{PublicationEvent, MusicEvent, ScreeningEvent, ComedyEvent, TheaterEvent, EducationEvent, VisualArtsEvent, Other}.
For movie genres, the genre-specific datasets are denoted by the first three letters of the 
respective genre as follows \textit{\{Drama, Comedy, Action, Thriller, Romance, 
Documentary, Adventure\}} = \{\textit{Dra, Com, Act, Thr, Rom, Doc, Adv}\}.  
$Movies$ refers to average values for all genres. The sampling method is denoted by the subscript, where $s$ represents \textit{stratified random sampling} and $p$ \textit{pld-aware sampling}.

\begin{table}
	\caption{Overview of the dataset size and contained plds. Movie genres are
	abbreviated by their first three letters. An own dataset for each genre is extracted since each genre is treated as a binary classification problem.}
	\label{dat:DatasetNotation}
	\begin{tabular}{lrrrr}
		\hline
		\textbf{Dataset} & \textbf{Size} & \textbf{Distinct plds} & \textbf{Avg. Instances/pld}\\
		\hline
		
		$Events_s$ &	67,744 &	1,482 & 45.71 \\
		$Events_p$ &	67,744 &	2,064 & 32.82 \\
		$Dra_s$ & 239,030 & 360 & 663.97 \\
		$Dra_p$ & 239,030 & 476 & 502.16 \\
		$Com_s$ & 239,030 & 342 & 698.92 \\
		$Com_p$ & 239,030 & 476 & 502.16 \\
		$Act_s$ & 239,030 & 361 & 662.13 \\
		$Act_p$	& 239,030 & 476 & 502.16 \\
		$Thr_s$ & 239,030 & 342 & 698.92 \\
		$Thr_p$ & 239,030 & 476 & 502.16 \\
		$Rom_s$ & 239,030 & 347 & 688.85 \\
		$Rom_p$	& 239,030 & 476 & 502.16 \\
		$Doc_s$ & 239,030 & 337 & 709.29 \\
		$Doc_p$	& 239,030 & 476 & 502.16 \\
		$Adv_s$ & 239,030 & 340 & 703.03 \\
		$Adv_p$	& 239,030 & 476 & 502.16 \\
		
		$Movies_s$ & 239,030 & 347 & 689.30 \\
		$Movies_p$ & 239,030 & 476 & 502.16 \\				
		\hline		
	\end{tabular}

\end{table}


\subsection{Metrics}
To evaluate the performance of the different classifiers, we compute the
following metrics:

	\textbf{Precision}: The fraction of the correctly classified instances among
	the instances assigned to one class.
	
	\textbf{Recall}: The fraction of the correctly assigned instances among all
	instances of the class.
	
	\textbf{F1 score}: The harmonic mean of recall and precision.
This work considers the F1 score to be the most relevant metric since it reflects 
both recall and precision.

\subsection{Baselines}
\label{sec:baselines}

\label{sec:BaseConf}

We compare our approach to the following baselines:

\textbf{\sc \bf RANDOM}: This baseline chooses a class at random.

	\textbf{\sc \bf SD-TYPE}: This baseline leverages conditional probabilities to
	infer the subject types using the \emph{SD-Type} approach \cite{Paulheim:2013}.
	The probabilities are based on the incoming and outgoing statements of 
	a particular node. Since \emph{SD-Type} was not originally designed to be
	applied to Web markup, we adapted it by only considering outgoing statements. 
	This is motivated by the fact that a complete set of incoming statements 
	can not be obtained for Web markup, 
	where links might (but are unlikely to) originate from any Web page. 

	 \textbf{\sc \bf KG-B}: This baseline employs a knowledge graph to obtain class
	 labels. The \textit{s:name} of a subject is used as input 
	 for \emph{DBpedia Spotlight} \cite{Daiber:2013:IEA:2506182.2506198} to 
	 obtain candidate entities from \emph{DBpedia} (\textit{dbp}). 
	 If the markup is annotated in one of the 12 languages supported by 
	 Spotlight\footnote{\url{http://www.dbpedia-spotlight.org/faq}}, the corresponding 
	 Spotlight model is used. For all other cases we employ the English Spotlight 
	 model. 
	 Labels obtained from DBpedia may be different from labels found in Web markup 
	 (e.g. the genre of the movie "Forrest Gump" is 
	 stated to be \emph{Drama} and \emph{Comedy} in DBpedia, 
	 but marked as \emph{Drama} and \emph{Romance} on \emph{imdb.com}). 
	 In order to avoid noisy and costly matching process, 
	 we address this issue by considering all candidates with a 
	 confidence of at least 0.5 as true positives as long as the matching 
	 entity shows the correct type (\textit{dbp:Event or \textit{s:Event}} 
	 respectively \textit{dbp:Movie} or \textit{s:Movie}), independent of 
	 whether or not the entity actually shows the expected categorical property.
	 If no candidate with a suitable type is found, the instance is assigned to the "Other"-class. Note that this simplification significantly boosts the performance of this otherwise naive baseline, yet serves the purpose of illustrating the lack of sufficient coverage (Section \ref{sec:results}).

	\section{Results}
\label{sec:results}

This section presents the results on the classification performance, 
the influence of the sampling methods and the individual features. 

\subsection{Classification Performance}

\begin{table*}[t]
	\caption{Macro averages for precision, recall, and F1 score [\%] over all datasets.}
	\begin{tabular}{l@{\quad}rrr@{\hskip 1em}rrr@{\hskip 3em}rrr@{\hskip 1em}rrr}
		\toprule
		\multirow{2}{*}{\raisebox{-\heavyrulewidth}{\textbf{Classifier}}} & \multicolumn{3}{c}{$Events_s$} 
		& \multicolumn{3}{c}{$Events_p$} & \multicolumn{3}{c}{$Movies_s$} & \multicolumn{3}{c}{$Movies_p$} \\    
		\cmidrule(l{0pt}r{6pt}){2-4} \cmidrule(l{0pt}r{26pt}){5-7}   
		\cmidrule(l{0pt}r{8pt}){8-10} \cmidrule(l{0pt}r{4pt}){11-13}      
		& Precision & Recall & F1 &  Precision & Recall & F1  &  Precision & Recall & F1  &  Precision & Recall & F1 \\

		\midrule
		\textsc{Random} & 12.72 & 12.71  &  12.71
						& 12.81 & 12.82 & 12.81 
						& 50.00 & 50.00 & 50.00 
						& 49.87 & 49.87 & 49.86  \\
		\textsc{SD-Type} & 58.35  & 49.56 &  40.98
						& 58.71 & 62.83 & 56.99 
						& 61.67 & 58.92 & 56.36 
						& 68.34 & 67.77 & 67.62 \\
						
		\textsc{KG-B}     & 39.06 & 12.59 & 02.96
						& 39.06 & 12.59 & 02.96
						& \textbf{76.52} & 55.70 & 44.82
						& 76.94 & 57.16 & 47.42 \\

		\midrule	
		\textsc{Na\"ive Bayes} & \textbf{86.04} & 44.24 & 40.04
							   & \textbf{84.06} & 50.51 & 47.78
							   & 69.06 & 50.29 & 33.98
							   & 61.55 & 50.39 & 34.19 \\
		
		\textsc{Decision Tree} & 70.60 & 70.26 & 70.15  
							   & 78.70 & 77.78 & 77.25
							   & 72.95 & 72.88 & 72.85 
							   & 82.01 & 81.89 & 81.86  \\ 
		
		\textsc{Random Forest} & 73.34 & \textbf{72.46} &\textbf{ 71.67} 
								& 80.75 & \textbf{79.71} & \textbf{79.59}
								& 74.62 & \textbf{74.49} & \textbf{74.46}
								& \textbf{83.27} & \textbf{83.16} & \textbf{83.14} \\

		\textsc{SVM}            & 75.51 & 70.10 & 67.64  
								& 81.45 & 78.67 & 77.34
								& 72.84 & 72.42 & 72.27
								& 81.75 & 81.37 & 81.27 \\
		\bottomrule
	\end{tabular}
	\label{ev:tab:Average}
\end{table*}

	Table \ref{ev:tab:Average} summarises the overall results of the baselines (\textsc{Random}, \textsc{SD-Type}, \textsc{KG-B}) as well as our proposed classification models (\textsc{Na\"ive Bayes}, \textsc{Decision Tree}, \textsc{Random Forest},	\textsc{SVM}).
	For both tasks, we report the macro averages of the results 
	with respect to precision, recall and F1 scores
	for both \emph{stratified random sampling} and \emph{pld-aware sampling}. 
 We observe that, for $Movies$, \textsc{Random Forest}, closely followed by \textsc{Decision Tree}, performs best across all evaluation metrics, except for precision/$Movies_s$, where it is slightly outperformed by \textsc{KG-B}. This is caused by the underlying assumption of the \textsc{KG-B} baseline that any entity match is considered as successful information inference, which unfairly boosts the baseline performance, in particular for popular entities. For $Events$, \textsc{Random Forest} shows the highest Recall and F1, closely followed by \textsc{Decision Tree}, whereas highest precision is achieved by \textsc{Na\"ive Bayes} in this case.
	The use of a single Decision Tree already results in relatively high F1 scores, e.g. 81.86\% for $Movies_p$.
	Considering a \textsc{Random Forest} as an ensemble of Decision Trees, we conclude that additional trees only slightly improve the outcome (F1 of 83.14\%). 
	The \textsc{SD-Type} baseline achieves F1 scores of 56.99\%  for $Events$. This significant difference in
	performance between the baseline and our approach reflects the
	fundamental difference between knowledge graphs and data sourced from markup and the need to consider features beyond the structural connections of entity descriptions when dealing with markup data. 
	For both \emph{Events} and \emph{Movies}, \textsc{KG-B}  assigns the vast majority of the instances to the "Other"-class, resulting in high recall and low precision for the aforementioned class. Due to the design of the baseline, all classes different from "Other" exhibit 100\% precision but very low recall, which ultimately results in low F1 scores after computing the macro average across classes. 

	For \emph{Movies}, Table \ref{ev:tab:Average} reports the average scores of the individual 
	genre-specific classifiers. It is worth to mention that the boundary of the classes (genres) might be fuzzy, e.g. it could be hard to differentiate a movie of genre "Thriller" from a movie of genre "Action". Since the classification of each genre 
	is formulated as a binary classification problem, the \textsc{random}-baseline performance is close to 50\% for all classes. The highest F1 score achieved by \textsc{SD-Type} is
	67.62\%, indicating that the subject properties used
	by this baseline might not be sufficient to classify movie genres precisely. Overall performance of the \textsc{KG-B} baseline is better in this task, driven by higher recall for instances of type movie, which are better represented in knowledge bases. Similar to our observations in the event classification task, \textsc{Random Forest} performs best, closely followed by \textsc{Decision Tree}. The F1 score of 83.14\% for \textsc{Random Forest} 
	significantly outperforms the baselines (paired t-test with $p$ < 0.01) when comparing \textsc{Random Forest} against the baselines in all configurations.
	Overall \textsc{Random Forest} classification using the features proposed in this paper clearly outperforms the baselines in both tasks.

\subsubsection{Classification Hyperparameter}
For each classifier used with an exception of the Na\"ive Bayes classifier, we determine the parameters that maximise the F1 score by employing the random search algorithm proposed by Bergstra and Bengio \cite{Bergstra:2012:RSH:2188385.2188395}. 
The Na\"ive Bayes classifier does not exhibit parameters that could be
optimised.
Table \ref{ev:tabParameters} gives an overview of the parameters that were considered during the optimisation, whereas Table \ref{ev:tabHyperParamterValues} summarises the hyper-parameters that were determined using random search. All previously shown performance results were obtained using the specified hyper-parameters.

\begin{table}
	\caption{Hyperparameters considered for optimisation.}
	\begin{tabular}{lll}
		\toprule
		\textbf{Classifier} & \textbf{Parameter} &\textbf{Range}\\
		\midrule
		{\sc Decision Tree} & Criterion & \makecell{Gini Impurity,\\Information Gain}\\
		& Min.Impurity Decrease & [0,1] \\
		\\
		{\sc Random Forest} & Criterion & \makecell{Gini Impurity,\\Information Gain}\\
		& Min.Impurity Decrease & [0,1] \\
		& No. Estimators & [5,20]\\
		\\
		{\sc SVM} & Penalty & [0,5] \\
		& Stopping Tolerance & [0,$10^{-3}$]\\ 
		\bottomrule
	\end{tabular}
	
	\label{ev:tabParameters}
\end{table}

\begin{table}
	\caption{Summary of classifier hyperparameters determined with random search for the following parameters: Crit: Criterion, Imp: Min. Impurity Decrease, No: No. Estimators, Pen: Penalty, Tol: Stopping Tolerance.}

	\begin{tabular}{lllllllc}
		\toprule
		\multirow{2}{*}{\raisebox{-\heavyrulewidth}{\textbf{Dataset}}} & \multicolumn{2}{c}{\textsc{Decision Tree}} & \multicolumn{3}{c}{\textsc{Random Forest}} & \multicolumn{2}{c}{\textsc{SVM}} \\

		\cmidrule(l{4pt}r{4pt}){2-3} \cmidrule(l{4pt}r{4pt}){4-6}\cmidrule(l{4pt}r{4pt}){7-8}
		
		& \textbf{Crit} & \textbf{Imp} & \textbf{Crit} &\textbf{Imp} &  \textbf{No} & \textbf{Pen} & \textbf{Tol} \\
		\midrule
		$Events_s$ & ent. & 0.192 & ent. & 0.892 & 13  & 3.53 & 0.0043 \\
		$Events_p$ & gini & 0.527 & ent. & 0.892 & 13 & 1.88 & 0.0098 \\
		$Dra_s$  & gini & 0.360 &  ent. & 0.938 & 16 & 0.66 & 0.0037 \\
		$Dra_p$  & gini & 0.360 & gini & 0.414 & 18 & 0.66 & 0.0037 \\
		$Com_s$  & gini & 0.360 & gini & 0.414 & 18 & 0.66 & 0.0037 \\
		$Com_p$  & ent. & 0.608 & gini & 0.160 & 20 & 0.66 & 0.0037 \\
		$Act_s$  & gini & 0.360 & gini & 0.160 & 20 & 0.66 & 0.0037 \\
		$Act_p$  & ent. & 0.558 & gini & 0.160 & 20 & 0.66 & 0.0037 \\
		$Thr_s$  & gini & 0.360 & ent. & 0.482 & 16 & 0.66 & 0.0037 \\
		$Thr_p$  & ent. & 0.608 & ent. & 0.482 & 13 & 0.66 & 0.0037 \\
		$Rom_s$  & ent. & 0.608 & ent. & 0.482 & 13 & 0.66 & 0.0037 \\
		$Rom_p$  & ent. & 0.287 & gini & 0.160 & 20 & 0.66 & 0.0037 \\
		$Doc_s$  & ent. & 0.192 & gini & 0.160 & 20 & 0.66 & 0.0037 \\
		$Doc_p$  & ent. & 0.099 & gini & 0.068 & 16 & 0.66 & 0.0037 \\
		$Adv_s$  & ent. & 0.287 & gini & 0.160 & 20 & 0.66 & 0.0037 \\
		$Adv_p$  & ent. & 0.287 & gini & 0.068 & 16 & 0.66 & 0.0037\\
		\bottomrule
		
	\end{tabular}
	
	\label{ev:tabHyperParamterValues}
\end{table}

\subsection{Influence of Sampling Methods}
In this section, we discuss the influence of the different sampling methods. 
Since the \textsc{Random Forest} classifier achieves the best results, we investigate the effects of sampling methods on our \textsc{Random Forest} configuration. 
\pgfplotstableread[col sep=comma,]{figures/samplingComp.csv}\samplingComp

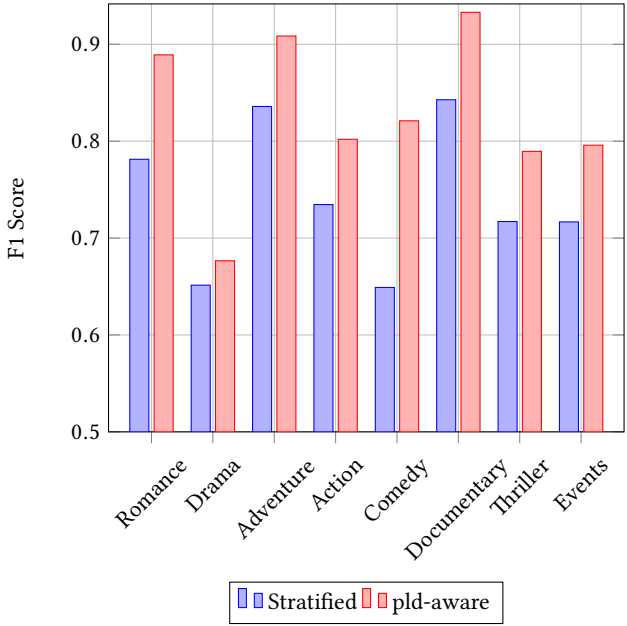
\begin{figure}
	\begin{tikzpicture}
	\begin{axis}[ybar,
	ymin=0.5,
	xtick=data,
	xticklabel style={rotate=45},
	xticklabels from table={\samplingComp}{Dataset},
	ylabel={F1 Score},
	xtick pos=left,  
	legend style={at={(0.5,-0.35)}, anchor=north,legend columns=-1},      
	grid=major,	
	enlarge y limits={0.02,upper},    
	bar width=0.8em]
	\addplot table[x expr=\coordindex, y={Stratified}]{\samplingComp};
	\addplot table[x expr=\coordindex, y={tld/pld}]{\samplingComp};
	\legend{Stratified, pld-aware}
	\end{axis}
	\end{tikzpicture}
	\caption{F1 scores macro averages [\%] for the \textsc{Random Forest} classifier with respect to dataset and sampling method.}
	\label{res:samplingComp}
\end{figure}

Figure \ref{res:samplingComp} shows the F1 scores with respect to the sampling method for $Events$ and the individual $Movies$ genre datasets. The use of \emph{pld-aware sampling} yields 
up to 17\% percentage points better results than the use of \emph{stratified random sampling}. 

We observe that the use of a more diverse training set 
(i.e. a dataset including more data from long-tail domains e.g. obtained through
the \emph{pld-aware sampling}) has a significant and beneficial effect on the classification outcome (paired
t-test with $p$ < 0.03).

\subsection{Influence of Features}	
In this section, we discuss the influence of the proposed features. 
We focus on the best performing classifier (\textsc{Random Forest}) while investigating the effects of varying the feature set.

	\begin{table}
		\caption{Random Forest F1 scores macro averages [\%] for different feature combinations (\emph{Events} datasets).}
		\begin{tabular}{lll}
			\toprule
			\textbf{Features} & \textbf{\emph{Events}$_{\textit{s}}$}  &	\textbf{\emph{Events}$_{\textit{p}}$} \\
			\midrule
			\emph{tld/pld} & 65.30 & 76.29	 \\
			\emph{page-vocab} & 62.57  & 79.8 \\
			\emph{node-vocab} & 60.66 & 68.09	 \\
			\emph{tld/pld,page-vocab} &	71.01 &	80.03 \\
			\emph{tld/pld,node-vocab} & 65.38 & 77.65 \\
			\emph{page-vocab,node-vocab} & 71.70 & 80.27	 \\
			\emph{tld/pld,page-vocab,node-vocab} & 71.67 & 79.59 \\
			\bottomrule
		\end{tabular}
		\label{res:EventFeatures}
	\end{table}
	
		\begin{table}
			\caption{Random Forest F1 scores macro averages [\%] for different feature combinations (\emph{Movies} datasets).}
			\begin{tabular}{lll}
				\toprule
				\textbf{Features} & \textbf{\emph{Movies}$_{\textit{s}}$}  &	\textbf{\emph{Movies}$_\textit{{p}}$} \\
				\midrule
				\emph{tld/pld} & 66.32  & 80.56	\\
				\emph{page-vocab} & 72.27  & 82.14  \\
				\emph{node-vocab} & 73.96 & 82.18\\
				\emph{tld/pld,page-vocab} & 72.59 &	82.68 \\
				\emph{tld/pld,node-vocab} &	74.28 & 82.94	\\
				\emph{page-vocab,node-vocab} & 74.33 & 82.59 \\
				\emph{tld/pld,page-vocab,node-vocab} & 74.46  & 83.14  \\
				\bottomrule
			\end{tabular}
			\label{res:MovieFeatures}
		\end{table}
	
	Table \ref{res:EventFeatures} presents the F1 scores obtained through \textsc{Random
	Forest} on the $Events$ dataset with respect to different feature combinations.
	Our results indicate that the influence of features varies strongly dependent
	on the respective types and classes. This seems intuitive, given that some
	classes might be more specifically characterised by certain features, 
	such as a set of plds.
	The \emph{tld/pld} features alone result in a reasonable performance 
	for \textit{Events} but not for \textit{Movies}. This indicates that the 
	source of the markup node is stronger correlated with its actual type or
	category for $Events$ than for $Movies$. This seems intuitive, 
	given that event-centred Websites tend to be more focused on certain 
	event types than movie-centred Websites are focused on particular genres. 
	However, these observations are likely to vary strongly dependent on the actual
	classification task. In contrast, the \emph{node-vocab} alone is not 
	sufficient to determine the event subtype with high F1 score.
	This observation corresponds to the insufficient performance of the
	\emph{SD-Type} baseline.

	The combination of \emph{tld/pld} and \emph{node-vocab} results only in a 
	slight improvement 
	of the results for $Events_p$. A dependence between the two features seems 
	intuitive 
	as pages extracted from the same pld are likely to be maintained by
	the same organisation and thus typically use the same set of \emph{schema.org}
	terms. For instance, an event database is likely to assign the same set of properties to each event resulting in a characteristic \emph{node-vocabulary} for the events of a single pld.
	Since the \emph{page-vocab} considers the terms that occur on the whole page, the number of considered terms is higher, which results in better chances to find usage of the same terms on other Web pages. This is reflected by the fact that both combinations of \emph{tld/pld, page-vocab} and \emph{page-vocab, node-vocab} lead to an improvement while the performance of \emph{tld/pld, node-vocab} is roughly the same as \emph{tld/pld} only. The combination of all three features yields in a slight decrease of the F1 
	score compared to \emph{tld/pld, page-vocab} only, indicating once more that the information contributed by \emph{node-vocab} is already provided by \emph{tld/pld}.
	
	Table \ref{res:MovieFeatures} shows average F1 scores using the \textsc{Random
	Forest} classifier on the $Movies$ datasets. In contrast to the $Events$ datasets 
	we can achieve relatively good performance by employing 
	only the \emph{node-vocab} feature. 
	Another difference is that we can observe a slightly larger margin between the exclusive use of \emph{node-vocab} and \emph{tld/pld}. This indicates that markup of movies of certain genres tend to exhibit the same \emph{schema.org} terms. 
	Any combination of two or more features results in similar 
	outcomes (with approximatly 1 percentage point difference). 
	In both domains we can see a substantial difference in the performance with respect to the sampling methods for all feature combinations. \emph{pld-aware sampling} consistently achieves higher F1 scores than \emph{stratified random sampling}, leading to the conclusion that individual features and feature combinations benefit from \emph{pld-aware sampling}.

\subsection{Discussion}
\label{sec:discussion}
Our experiments illustrated that traditional knowledge graph completion approaches that are not specifically designed for Web markup data may not be directly applicable to this kind of data, mainly due to the sparsity of individuals and the lack of connectivity in Web markup. Moreover, we observed that it is not sufficient to consider only node-specific features such as \emph{node-vocab} to infer missing categorical information in Web markup. In contrast, contextual features such as \emph{tld/pld} and \emph{page-vocab} provide important information to infer missing statements.

In particular, our experiments demonstrated that contextual
features such as \emph{tld/pld} and \emph{page-vocab} are discriminative for both tasks under consideration. These features are effective because many Websites focus on a particular topic, e.g. theater or music events. 
We observed that the \emph{page-vocab} feature is especially useful in both tasks, as it describes the context of the particular node in a more specific way.  
Whereas the use of the \emph{tld/pld} feature can naturally only be applied to instances from known plds, i.e. plds that are contained in the training data, performance drops are expected when classifying data from unknown plds. However, our results indicate that features representative for certain kinds of plds, such as \emph{page-vocab}, can serve as potent substitute able to efficiently classify markup from unknown sources.

Limitations arise from the focus on two particular tasks only. We anticipate variation in performance of particular features when applying this approach to other kinds of categorical information. Similarly, considering that our ground truth has been constructed by relying on markup nodes where the sought-after information was present already on the Web, one might argue that this constraint has led to a bias towards markup nodes of generally higher quality. Additional experiments on an unconstrained and randomly selected ground truth will investigate this assumption further as part of future work.

	\vspace{-6pt}
	
	\section{Related Work}
\label{sec:related}

In this section we discuss related work in the areas of knowledge
graph completion and schema inference for traditional knowledge graphs along with works focused directly on Web markup.

Existing approaches to knowledge graph completion
and dataset profiling including its applications
to schema inference have been summarised in 
recent survey articles \cite{paulheim2016knowledge, benellefi2017dataset}.
These approaches include in particular entity type inference, 
relation prediction and relation validation.
In the context of KG completion, entity type inference is
most commonly addressed as a multi-class prediction problem. 
\cite{Paulheim:2013} makes use of properties and 
conditional probabilities to infer entity types, building the baseline for our
approach.
Schemex is an approach to extract and index schema information from 
Linked Open Data (LOD) \cite{Konrath:2012:SEC:2399444.2399563}.
In YAGO+F instance-based matching enables to enrich Freebase entities with
YAGO concepts \cite{yago+f}.
\cite{Gottron2013} made use of Schemex to analyse schema information of 
LOD and found that properties provide information about subject types. 
In our work, we use properties as features for inferring missing categorical
information in general.
\cite{DBLP:conf/emnlp/GardnerM15} predicts relations between two nodes by 
leveraging random walk inference methods using sub-graphs to improve the path
ranking algorithm (PRA), initially proposed in
\cite{Lao:2010:RRU:1842816.1842823}.
\cite{knowledge-base-completion-via-coupled-path-ranking} also builds on PRA and extends 
it to a multi-task learning approach. 

All of the works discussed above have been applied to traditional 
KGs such as DBpedia, NELL and YAGO. In contrast, in
this work we aim at inferring information on the Web markup data. Web markup is distinguished from the aforementioned knowledge graphs by specific characteristics, i.e. annotations are often very sparse or noisy, vocabularies
are not used correctly in many cases and the overall RDF graph is connected very loosely \cite{Dietze:2017:AIE:3041021.3054160,
Meusel:2014:WMR:2717213.2717235}.
For these reasons, existing KG completion methods are not likely to perform well on Web markup. For instance, KG completion approaches based on graph topology (e.g. relation
prediction discussed above) rely on the presence of relations, which are not widely available in markup.

Various approaches employ embeddings for KG completion in traditional knowledge graphs.
\cite{DBLP:journals/tkde/WangMWG17} conducted a survey on KG embeddings for 
applications such as link prediction, entity classification and triple classification. 
\cite{Wang:2015:KBC:2832415.2832507} makes use of embeddings and rules.
\cite{Lin:2015:LER:2886521.2886624} propose the \emph{TransR} model that builds 
separate entity and relation embeddings to compute the plausibility of
missing triples. \cite{NIPS2013_5028} predicts relations between entities 
by employing neural tensor networks.
Embeddings techniques have not yet been applied to Web markup yet lend themselves as direction for future research.

Several recent studies focused on analysing the characteristics, evolution and coverage of markup \citep{2016lileAnalysing, sahoo2015analysing, Dietze:2017:AIE:3041021.3054160} and on addressing specific tasks in the context of Web markup. Meusel et al. proposed heuristics that can be employed to fix common errors in Web markup \cite{Meusel:2014:WMR:2717213.2717235, conf/esws/MeuselP15}. In this work, we applied the heuristics proposed in \cite{conf/esws/MeuselP15} for pre-processing and data cleansing. 
\cite{conf/icde/YuGFD17, yu2016towards} provide pipelines for data fusion and entity summarisation on Web markup, involving heuristics, clustering and supervised approaches for entity matching and classification of markup statements. \cite{yu2018knowmore} builds on these works by utilising fused markup data to augment existing knowledge bases, showing the complementarity of markup data and its potential to significantly complement information from traditional reference KGs. 

While these works demonstrate the use of markup data, 
they suffer from the sparsity of individual nodes.  
The inference approach proposed in our work can augment markup nodes and is
likely to boost the performance on both fusion as well as KG augmentation tasks.
In particular, considering the impact of the use of controlled vocabularies 
on data reuse \cite{endris2017dataset}, we anticipate that inference of 
crucial categorical information can facilitate reuse of markup data.

	\vspace{-6pt}
	
	\section{Conclusion \& Future Work}
\label{sec:conclusion}

In this work, we addressed the problem of interpreting noisy and sparse Web markup by proposing an approach for automatically inferring categorical information for particular entities, thereby augmenting sparse markup nodes with information, which often is 
essential when interpreting markup and the corresponding Web pages.
We leveraged the large amount of publicly available data as training data for a supervised machine learning approach. We employed Web markup specific features 
such as \emph{tld/pld}, \emph{node vocabulary} and \emph{page vocabulary} and conducted an extensive evaluation of different classification algorithms, sampling methods and feature sets. Our proposed configuration outperforms existing baselines significantly, with \textsc{Random Forest} providing the most consistent performance across classes and datasets.

By applying our approach to the problem of inferring event types and movie genres,  we demonstrated that supervised inference can uncover entity-centric categorical information, which is essential when interpreting markup or Websites in general. Potential applications include knowledge base augmentation from Web markup \cite{yu2018knowmore}, Website classification or Web search in general.
Considering the still limited experiments and the limitations of our dataset (Section \ref{sec:discussion}), future work will include the conduction of additional experiments, involving more diverse datasets and tasks. With respect to the latter, current experiments are being conducted on discretised information rather than properties, which are apriori categorical. In addition, we aim at investigating the impact of our approach when being included into data fusion and KG augmentation pipelines, such as \cite{conf/icde/YuGFD17, yu2018knowmore}.

\footnotesize{
	\begin{acks}
	This work was partially funded by the European Commission ("AFEL" project, grant ID 687916) and the BMBF ("Data4UrbanMobility" project, grant ID 02K15A040).
	\end{acks}
	}

	\balance
	\bibliographystyle{ACM-Reference-Format}
		\bibliography{references}


\begin{thebibliography}{00}


\ifx \showCODEN    \undefined \def \showCODEN     #1{\unskip}     \fi
\ifx \showDOI      \undefined \def \showDOI       #1{#1}\fi
\ifx \showISBNx    \undefined \def \showISBNx     #1{\unskip}     \fi
\ifx \showISBNxiii \undefined \def \showISBNxiii  #1{\unskip}     \fi
\ifx \showISSN     \undefined \def \showISSN      #1{\unskip}     \fi
\ifx \showLCCN     \undefined \def \showLCCN      #1{\unskip}     \fi
\ifx \shownote     \undefined \def \shownote      #1{#1}          \fi
\ifx \showarticletitle \undefined \def \showarticletitle #1{#1}   \fi
\ifx \showURL      \undefined \def \showURL       {\relax}        \fi
\providecommand\bibfield[2]{#2}
\providecommand\bibinfo[2]{#2}
\providecommand\natexlab[1]{#1}
\providecommand\showeprint[2][]{arXiv:#2}

\bibitem[\protect\citeauthoryear{Ben~Ellefi, Bellahsene, John, Demidova,
  Dietze, Szymanski, and Todorov}{Ben~Ellefi et~al\mbox{.}}{2017}]%
        {benellefi2017dataset}
\bibfield{author}{\bibinfo{person}{Mohamed Ben~Ellefi}, \bibinfo{person}{Zohra
  Bellahsene}, \bibinfo{person}{Breslin John}, \bibinfo{person}{Elena
  Demidova}, \bibinfo{person}{Stefan Dietze}, \bibinfo{person}{Julian
  Szymanski}, {and} \bibinfo{person}{Konstantin Todorov}.}
  \bibinfo{year}{2017}\natexlab{}.
\newblock \showarticletitle{RDF Dataset Profiling - a Survey of Features,
  Methods, Vocabularies and Applications}.
\newblock \bibinfo{journal}{{\em Semantic Web Journal\/}}
  (\bibinfo{year}{2017}).
\newblock
\newblock
\shownote{to appear.}


\bibitem[\protect\citeauthoryear{Bergstra and Bengio}{Bergstra and
  Bengio}{2012}]%
        {Bergstra:2012:RSH:2188385.2188395}
\bibfield{author}{\bibinfo{person}{James Bergstra} {and}
  \bibinfo{person}{Yoshua Bengio}.} \bibinfo{year}{2012}\natexlab{}.
\newblock \showarticletitle{Random Search for Hyper-parameter Optimization}.
\newblock \bibinfo{journal}{{\em J. Mach. Learn. Res.\/}}  \bibinfo{volume}{13}
  (\bibinfo{date}{Feb.} \bibinfo{year}{2012}), \bibinfo{pages}{281--305}.
\newblock
\showISSN{1532-4435}


\bibitem[\protect\citeauthoryear{Bizer, Eckert, Meusel, M{\"u}hleisen,
  Schuhmacher, and V{\"o}lker}{Bizer et~al\mbox{.}}{2013}]%
        {Bizer2013}
\bibfield{author}{\bibinfo{person}{Christian Bizer}, \bibinfo{person}{Kai
  Eckert}, \bibinfo{person}{Robert Meusel}, \bibinfo{person}{Hannes
  M{\"u}hleisen}, \bibinfo{person}{Michael Schuhmacher}, {and}
  \bibinfo{person}{Johanna V{\"o}lker}.} \bibinfo{year}{2013}\natexlab{}.
\newblock \bibinfo{booktitle}{{\em Deployment of RDFa, Microdata, and
  Microformats on the Web -- A Quantitative Analysis}}.
\newblock \bibinfo{publisher}{Springer Berlin Heidelberg},
  \bibinfo{pages}{17--32}.
\newblock


\bibitem[\protect\citeauthoryear{Bollacker, Evans, Paritosh, Sturge, and
  Taylor}{Bollacker et~al\mbox{.}}{2008}]%
        {Bollacker:2008:FCC:1376616.1376746}
\bibfield{author}{\bibinfo{person}{Kurt Bollacker}, \bibinfo{person}{Colin
  Evans}, \bibinfo{person}{Praveen Paritosh}, \bibinfo{person}{Tim Sturge},
  {and} \bibinfo{person}{Jamie Taylor}.} \bibinfo{year}{2008}\natexlab{}.
\newblock \showarticletitle{Freebase: A Collaboratively Created Graph Database
  for Structuring Human Knowledge}. In \bibinfo{booktitle}{{\em Proceedings of
  the 2008 ACM SIGMOD International Conference on Management of Data}} {\em
  (\bibinfo{series}{SIGMOD '08})}. \bibinfo{publisher}{ACM},
  \bibinfo{pages}{1247--1250}.
\newblock


\bibitem[\protect\citeauthoryear{Daiber, Jakob, Hokamp, and Mendes}{Daiber
  et~al\mbox{.}}{2013}]%
        {Daiber:2013:IEA:2506182.2506198}
\bibfield{author}{\bibinfo{person}{Joachim Daiber}, \bibinfo{person}{Max
  Jakob}, \bibinfo{person}{Chris Hokamp}, {and} \bibinfo{person}{Pablo~N.
  Mendes}.} \bibinfo{year}{2013}\natexlab{}.
\newblock \showarticletitle{Improving Efficiency and Accuracy in Multilingual
  Entity Extraction}. In \bibinfo{booktitle}{{\em Proceedings of the 9th
  International Conference on Semantic Systems}} {\em
  (\bibinfo{series}{I-SEMANTICS '13})}. \bibinfo{publisher}{ACM},
  \bibinfo{pages}{121--124}.
\newblock
\showISBNx{978-1-4503-1972-0}


\bibitem[\protect\citeauthoryear{Demidova, Oelze, and Nejdl}{Demidova
  et~al\mbox{.}}{2013}]%
        {yago+f}
\bibfield{author}{\bibinfo{person}{Elena Demidova}, \bibinfo{person}{Iryna
  Oelze}, {and} \bibinfo{person}{Wolfgang Nejdl}.}
  \bibinfo{year}{2013}\natexlab{}.
\newblock \showarticletitle{Aligning Freebase with the YAGO Ontology}. In
  \bibinfo{booktitle}{{\em Proceedings of the 22Nd ACM International Conference
  on Information \& Knowledge Management}} {\em (\bibinfo{series}{CIKM '13})}.
  \bibinfo{publisher}{ACM}, \bibinfo{pages}{579--588}.
\newblock


\bibitem[\protect\citeauthoryear{Dietze, Taibi, Yu, Barker, and d'Aquin}{Dietze
  et~al\mbox{.}}{2017}]%
        {Dietze:2017:AIE:3041021.3054160}
\bibfield{author}{\bibinfo{person}{Stefan Dietze}, \bibinfo{person}{Davide
  Taibi}, \bibinfo{person}{Ran Yu}, \bibinfo{person}{Phil Barker}, {and}
  \bibinfo{person}{Mathieu d'Aquin}.} \bibinfo{year}{2017}\natexlab{}.
\newblock \showarticletitle{Analysing and Improving Embedded Markup of Learning
  Resources on the Web}. In \bibinfo{booktitle}{{\em Proceedings of the 26th
  International Conference on World Wide Web Companion}} {\em
  (\bibinfo{series}{WWW '17 Companion})}. \bibinfo{publisher}{International
  World Wide Web Conferences Steering Committee}, \bibinfo{pages}{283--292}.
\newblock
\showISBNx{978-1-4503-4914-7}


\bibitem[\protect\citeauthoryear{Endris, Giménez-García, Thakkar, Demidova,
  Zimmermann, Lange, and Simperl}{Endris et~al\mbox{.}}{2017}]%
        {endris2017dataset}
\bibfield{author}{\bibinfo{person}{Kemele~M. Endris}, \bibinfo{person}{José~M.
  Giménez-García}, \bibinfo{person}{Harsh Thakkar}, \bibinfo{person}{Elena
  Demidova}, \bibinfo{person}{Antoine Zimmermann}, \bibinfo{person}{Christoph
  Lange}, {and} \bibinfo{person}{Elena Simperl}.}
  \bibinfo{year}{2017}\natexlab{}.
\newblock \showarticletitle{Dataset Reuse: An Analysis of References in
  Community Discussions, Publications and Data}. In \bibinfo{booktitle}{{\em
  Proceedings of the Ninth International Conference on Knowledge Capture (K-CAP
  2017)}}.
\newblock


\bibitem[\protect\citeauthoryear{Gardner and Mitchell}{Gardner and
  Mitchell}{2015}]%
        {DBLP:conf/emnlp/GardnerM15}
\bibfield{author}{\bibinfo{person}{Matt Gardner} {and} \bibinfo{person}{Tom~M.
  Mitchell}.} \bibinfo{year}{2015}\natexlab{}.
\newblock \showarticletitle{Efficient and Expressive Knowledge Base Completion
  Using Subgraph Feature Extraction}. In \bibinfo{booktitle}{{\em Proceedings
  of the 2015 Conference on Empirical Methods in Natural Language Processing,
  {EMNLP} 2015}}. \bibinfo{pages}{1488--1498}.
\newblock


\bibitem[\protect\citeauthoryear{Gottron, Knauf, Scheglmann, and
  Scherp}{Gottron et~al\mbox{.}}{2013}]%
        {Gottron2013}
\bibfield{author}{\bibinfo{person}{Thomas Gottron}, \bibinfo{person}{Malte
  Knauf}, \bibinfo{person}{Stefan Scheglmann}, {and} \bibinfo{person}{Ansgar
  Scherp}.} \bibinfo{year}{2013}\natexlab{}.
\newblock \showarticletitle{A Systematic Investigation of Explicit and Implicit
  Schema Information on the Linked Open Data Cloud}. In
  \bibinfo{booktitle}{{\em Proceedings of the ESWC 2013}}.
  \bibinfo{publisher}{Springer Berlin Heidelberg}, \bibinfo{pages}{228--242}.
\newblock


\bibitem[\protect\citeauthoryear{Guha, Brickley, and Macbeth}{Guha
  et~al\mbox{.}}{2016}]%
        {Guha:2016:SES:2886013.2844544}
\bibfield{author}{\bibinfo{person}{R.~V. Guha}, \bibinfo{person}{Dan Brickley},
  {and} \bibinfo{person}{Steve Macbeth}.} \bibinfo{year}{2016}\natexlab{}.
\newblock \showarticletitle{Schema.Org: Evolution of Structured Data on the
  Web}.
\newblock \bibinfo{journal}{{\em Commun. ACM\/}} \bibinfo{volume}{59},
  \bibinfo{number}{2} (\bibinfo{date}{Jan.} \bibinfo{year}{2016}),
  \bibinfo{pages}{44--51}.
\newblock
\showISSN{0001-0782}


\bibitem[\protect\citeauthoryear{Konrath, Gottron, Staab, and Scherp}{Konrath
  et~al\mbox{.}}{2012}]%
        {Konrath:2012:SEC:2399444.2399563}
\bibfield{author}{\bibinfo{person}{Mathias Konrath}, \bibinfo{person}{Thomas
  Gottron}, \bibinfo{person}{Steffen Staab}, {and} \bibinfo{person}{Ansgar
  Scherp}.} \bibinfo{year}{2012}\natexlab{}.
\newblock \showarticletitle{SchemEX - Efficient Construction of a Data
  Catalogue by Stream-based Indexing of Linked Data}.
\newblock \bibinfo{journal}{{\em Web Semant.\/}}  \bibinfo{volume}{16}
  (\bibinfo{date}{Nov.} \bibinfo{year}{2012}), \bibinfo{pages}{52--58}.
\newblock
\showISSN{1570-8268}


\bibitem[\protect\citeauthoryear{Lao and Cohen}{Lao and Cohen}{2010}]%
        {Lao:2010:RRU:1842816.1842823}
\bibfield{author}{\bibinfo{person}{Ni Lao} {and} \bibinfo{person}{William~W.
  Cohen}.} \bibinfo{year}{2010}\natexlab{}.
\newblock \showarticletitle{Relational Retrieval Using a Combination of
  Path-constrained Random Walks}.
\newblock \bibinfo{journal}{{\em Mach. Learn.\/}} \bibinfo{volume}{81},
  \bibinfo{number}{1} (\bibinfo{date}{Oct.} \bibinfo{year}{2010}),
  \bibinfo{pages}{53--67}.
\newblock
\showISSN{0885-6125}


\bibitem[\protect\citeauthoryear{Lin, Liu, Sun, Liu, and Zhu}{Lin
  et~al\mbox{.}}{2015}]%
        {Lin:2015:LER:2886521.2886624}
\bibfield{author}{\bibinfo{person}{Yankai Lin}, \bibinfo{person}{Zhiyuan Liu},
  \bibinfo{person}{Maosong Sun}, \bibinfo{person}{Yang Liu}, {and}
  \bibinfo{person}{Xuan Zhu}.} \bibinfo{year}{2015}\natexlab{}.
\newblock \showarticletitle{Learning Entity and Relation Embeddings for
  Knowledge Graph Completion}. In \bibinfo{booktitle}{{\em Proceedings of the
  Twenty-Ninth AAAI Conference on Artificial Intelligence}} {\em
  (\bibinfo{series}{AAAI'15})}. \bibinfo{publisher}{AAAI Press},
  \bibinfo{pages}{2181--2187}.
\newblock
\showISBNx{0-262-51129-0}


\bibitem[\protect\citeauthoryear{Meusel and Paulheim}{Meusel and
  Paulheim}{2015}]%
        {conf/esws/MeuselP15}
\bibfield{author}{\bibinfo{person}{Robert Meusel} {and} \bibinfo{person}{Heiko
  Paulheim}.} \bibinfo{year}{2015}\natexlab{}.
\newblock \showarticletitle{Heuristics for Fixing Common Errors in Deployed
  Schema.Org Microdata}. In \bibinfo{booktitle}{{\em Proceedings of the 12th
  European Semantic Web Conference on The Semantic Web}}.
  \bibinfo{publisher}{Springer-Verlag New York, Inc.},
  \bibinfo{pages}{152--168}.
\newblock


\bibitem[\protect\citeauthoryear{Meusel, Petrovski, and Bizer}{Meusel
  et~al\mbox{.}}{2014}]%
        {Meusel:2014:WMR:2717213.2717235}
\bibfield{author}{\bibinfo{person}{Robert Meusel}, \bibinfo{person}{Petar
  Petrovski}, {and} \bibinfo{person}{Christian Bizer}.}
  \bibinfo{year}{2014}\natexlab{}.
\newblock \showarticletitle{The WebDataCommons Microdata, RDFa and Microformat
  Dataset Series}. In \bibinfo{booktitle}{{\em Proceedings of the 13th
  International Semantic Web Conference - Part I}} {\em (\bibinfo{series}{ISWC
  '14})}. \bibinfo{publisher}{Springer-Verlag New York, Inc.},
  \bibinfo{pages}{277--292}.
\newblock
\showISBNx{978-3-319-11963-2}


\bibitem[\protect\citeauthoryear{Meusel, Ritze, and Paulheim}{Meusel
  et~al\mbox{.}}{2016}]%
        {meusel2016towards}
\bibfield{author}{\bibinfo{person}{Robert Meusel}, \bibinfo{person}{Dominique
  Ritze}, {and} \bibinfo{person}{Heiko Paulheim}.}
  \bibinfo{year}{2016}\natexlab{}.
\newblock \showarticletitle{Towards More Accurate Statistical Profiling of
  Deployed schema.org Microdata}.
\newblock \bibinfo{journal}{{\em ACM Journal of Data and Information
  Quality\/}} \bibinfo{volume}{8}, \bibinfo{number}{1} (\bibinfo{year}{2016}).
\newblock


\bibitem[\protect\citeauthoryear{Paulheim}{Paulheim}{2016}]%
        {paulheim2016knowledge}
\bibfield{author}{\bibinfo{person}{Heiko Paulheim}.}
  \bibinfo{year}{2016}\natexlab{}.
\newblock \showarticletitle{Knowledge graph refinement: A survey of approaches
  and evaluation methods}.
\newblock \bibinfo{journal}{{\em Semantic Web\/}} \bibinfo{number}{Preprint}
  (\bibinfo{year}{2016}), \bibinfo{pages}{1--20}.
\newblock


\bibitem[\protect\citeauthoryear{Paulheim and Bizer}{Paulheim and
  Bizer}{2013}]%
        {Paulheim:2013}
\bibfield{author}{\bibinfo{person}{Heiko Paulheim} {and}
  \bibinfo{person}{Christian Bizer}.} \bibinfo{year}{2013}\natexlab{}.
\newblock \showarticletitle{Type Inference on Noisy RDF Data}. In
  \bibinfo{booktitle}{{\em Proceedings of the 12th International Semantic Web
  Conference - Part I}} {\em (\bibinfo{series}{ISWC '13})}.
  \bibinfo{publisher}{Springer-Verlag New York, Inc.},
  \bibinfo{pages}{510--525}.
\newblock
\showISBNx{978-3-642-41334-6}


\bibitem[\protect\citeauthoryear{Sahoo, Gadiraju, Yu, Saha, and Dietze}{Sahoo
  et~al\mbox{.}}{2016}]%
        {sahoo2015analysing}
\bibfield{author}{\bibinfo{person}{Pracheta Sahoo}, \bibinfo{person}{Ujwal
  Gadiraju}, \bibinfo{person}{Ran Yu}, \bibinfo{person}{Sriparna Saha}, {and}
  \bibinfo{person}{Stefan Dietze}.} \bibinfo{year}{2016}\natexlab{}.
\newblock \showarticletitle{Analysing Structured Scholarly Data Embedded in Web
  Pages}.
\newblock  (\bibinfo{date}{April} \bibinfo{year}{2016}).
\newblock


\bibitem[\protect\citeauthoryear{Singhal}{Singhal}{2012}]%
        {googleGraph}
\bibfield{author}{\bibinfo{person}{Amit Singhal}.}
  \bibinfo{year}{2012}\natexlab{}.
\newblock \bibinfo{title}{Introducing the Knowledge Graph: things, not
  strings}.
\newblock \bibinfo{howpublished}{Official Google Blog}.   (\bibinfo{date}{May}
  \bibinfo{year}{2012}).
\newblock
\showURL{%
\url{https://googleblog.blogspot.de/2012/05/introducing-knowledge-graph-things-not.html}}
\newblock
\shownote{accessed on 01/20/2018.}


\bibitem[\protect\citeauthoryear{Socher, Chen, Manning, and Ng}{Socher
  et~al\mbox{.}}{2013}]%
        {NIPS2013_5028}
\bibfield{author}{\bibinfo{person}{Richard Socher}, \bibinfo{person}{Danqi
  Chen}, \bibinfo{person}{Christopher~D Manning}, {and} \bibinfo{person}{Andrew
  Ng}.} \bibinfo{year}{2013}\natexlab{}.
\newblock \showarticletitle{Reasoning With Neural Tensor Networks for Knowledge
  Base Completion}.
\newblock In \bibinfo{booktitle}{{\em Advances in Neural Information Processing
  Systems 26}}. \bibinfo{publisher}{Curran Associates, Inc.},
  \bibinfo{pages}{926--934}.
\newblock


\bibitem[\protect\citeauthoryear{Suchanek, Kasneci, and Weikum}{Suchanek
  et~al\mbox{.}}{2007}]%
        {suchanek2007yago}
\bibfield{author}{\bibinfo{person}{Fabian~M. Suchanek},
  \bibinfo{person}{Gjergji Kasneci}, {and} \bibinfo{person}{Gerhard Weikum}.}
  \bibinfo{year}{2007}\natexlab{}.
\newblock \showarticletitle{Yago: A Core of Semantic Knowledge}. In
  \bibinfo{booktitle}{{\em Proceedings of the 16th International Conference on
  World Wide Web}} {\em (\bibinfo{series}{WWW '07})}. \bibinfo{publisher}{ACM},
  \bibinfo{pages}{697--706}.
\newblock


\bibitem[\protect\citeauthoryear{Taibi and Dietze}{Taibi and Dietze}{2016}]%
        {2016lileAnalysing}
\bibfield{author}{\bibinfo{person}{Davide Taibi} {and} \bibinfo{person}{Stefan
  Dietze}.} \bibinfo{year}{2016}\natexlab{}.
\newblock \showarticletitle{Towards embedded markup of learning resources on
  the Web: a quantitative Analysis of LRMI Terms Usage}. In
  \bibinfo{booktitle}{{\em Proceedings of the WWW Companion 2016}}.
\newblock


\bibitem[\protect\citeauthoryear{Wang, Liu, Luo, Wang, and Lin}{Wang
  et~al\mbox{.}}{2016}]%
        {knowledge-base-completion-via-coupled-path-ranking}
\bibfield{author}{\bibinfo{person}{Quan Wang}, \bibinfo{person}{Jing Liu},
  \bibinfo{person}{Yuanfei Luo}, \bibinfo{person}{Bin Wang}, {and}
  \bibinfo{person}{Chin-Yew Lin}.} \bibinfo{year}{2016}\natexlab{}.
\newblock \showarticletitle{Knowledge Base Completion via Coupled Path
  Ranking}. In \bibinfo{booktitle}{{\em Proceedings of the 54th Annual Meeting
  of the Association for Computational Linguistics (ACL)}}.
  \bibinfo{publisher}{ACL - Association for Computational Linguistics},
  \bibinfo{pages}{1308--1318}.
\newblock


\bibitem[\protect\citeauthoryear{Wang, Mao, Wang, and Guo}{Wang
  et~al\mbox{.}}{2017}]%
        {DBLP:journals/tkde/WangMWG17}
\bibfield{author}{\bibinfo{person}{Quan Wang}, \bibinfo{person}{Zhendong Mao},
  \bibinfo{person}{Bin Wang}, {and} \bibinfo{person}{Li Guo}.}
  \bibinfo{year}{2017}\natexlab{}.
\newblock \showarticletitle{Knowledge Graph Embedding: {A} Survey of Approaches
  and Applications}.
\newblock \bibinfo{journal}{{\em {IEEE} Trans. Knowl. Data Eng.\/}}
  \bibinfo{volume}{29}, \bibinfo{number}{12} (\bibinfo{year}{2017}),
  \bibinfo{pages}{2724--2743}.
\newblock


\bibitem[\protect\citeauthoryear{Wang, Wang, and Guo}{Wang
  et~al\mbox{.}}{2015}]%
        {Wang:2015:KBC:2832415.2832507}
\bibfield{author}{\bibinfo{person}{Quan Wang}, \bibinfo{person}{Bin Wang},
  {and} \bibinfo{person}{Li Guo}.} \bibinfo{year}{2015}\natexlab{}.
\newblock \showarticletitle{Knowledge Base Completion Using eddings and Rules}.
  In \bibinfo{booktitle}{{\em Proceedings of the 24th International Conference
  on Artificial Intelligence}} {\em (\bibinfo{series}{IJCAI'15})}.
  \bibinfo{publisher}{AAAI Press}, \bibinfo{pages}{1859--1865}.
\newblock
\showISBNx{978-1-57735-738-4}


\bibitem[\protect\citeauthoryear{Yu, Fetahu, Gadiraju, and Dietze}{Yu
  et~al\mbox{.}}{2016a}]%
        {yu2016iswc}
\bibfield{author}{\bibinfo{person}{Ran Yu}, \bibinfo{person}{Besnik Fetahu},
  \bibinfo{person}{Ujwal Gadiraju}, {and} \bibinfo{person}{Stefan Dietze}.}
  \bibinfo{year}{2016}\natexlab{a}.
\newblock \showarticletitle{A Survey on Challenges in Web Markup Data for
  Entity Retrieval}. In \bibinfo{booktitle}{{\em Proceedings of the {ISWC} 2016
  Posters {\&} Demonstrations Track}}.
\newblock


\bibitem[\protect\citeauthoryear{Yu, Gadiraju, Fetahu, and Dietze}{Yu
  et~al\mbox{.}}{2017a}]%
        {conf/icde/YuGFD17}
\bibfield{author}{\bibinfo{person}{Ran Yu}, \bibinfo{person}{Ujwal Gadiraju},
  \bibinfo{person}{Besnik Fetahu}, {and} \bibinfo{person}{Stefan Dietze}.}
  \bibinfo{year}{2017}\natexlab{a}.
\newblock \showarticletitle{FuseM: Query-Centric Data Fusion on Structured Web
  Markup}. In \bibinfo{booktitle}{{\em Proceedings of the IEEE 33rd
  International Conference on Data Engineering (ICDE), 2017}}.
  \bibinfo{publisher}{IEEE Computer Society}, \bibinfo{pages}{179--182}.
\newblock


\bibitem[\protect\citeauthoryear{Yu, Gadiraju, Fetahu, Lehmberg, Ritze, and
  Dietze}{Yu et~al\mbox{.}}{2017b}]%
        {yu2018knowmore}
\bibfield{author}{\bibinfo{person}{Ran Yu}, \bibinfo{person}{Ujwal Gadiraju},
  \bibinfo{person}{Besnik Fetahu}, \bibinfo{person}{Oliver Lehmberg},
  \bibinfo{person}{Dominique Ritze}, {and} \bibinfo{person}{Stefan Dietze}.}
  \bibinfo{year}{2017}\natexlab{b}.
\newblock \showarticletitle{KnowMore - Knowledge Base Augmentation with
  Structured Web Markup}.
\newblock \bibinfo{journal}{{\em Semantic Web Journal, IOS Press\/}}
  (\bibinfo{year}{2017}).
\newblock


\bibitem[\protect\citeauthoryear{Yu, Gadiraju, Zhu, Fetahu, and Dietze}{Yu
  et~al\mbox{.}}{2016b}]%
        {yu2016towards}
\bibfield{author}{\bibinfo{person}{Ran Yu}, \bibinfo{person}{Ujwal Gadiraju},
  \bibinfo{person}{Xiaofei Zhu}, \bibinfo{person}{Besnik Fetahu}, {and}
  \bibinfo{person}{Stefan Dietze}.} \bibinfo{year}{2016}\natexlab{b}.
\newblock \showarticletitle{Towards Entity Summarisation on Structured Web
  Markup}.
\newblock \bibinfo{journal}{{\em The Semantic Web: ESWC 2016 Satellite
  Events,\/}} (\bibinfo{date}{June} \bibinfo{year}{2016}).
\newblock


\end{thebibliography}
	\theendnotes

\end{document}